\documentclass[10pt,twocolumn,letterpaper]{article}

\usepackage{wacv}
\usepackage{times}
\usepackage{epsfig}
\usepackage{graphicx}
\usepackage{amsmath}
\usepackage{amssymb}

\usepackage{amsfonts}
\usepackage{bm}
\usepackage{mathtools}
\usepackage{pdfpages}

\usepackage[pagebackref=true,breaklinks=true,letterpaper=true,colorlinks,bookmarks=false]{hyperref}

\wacvfinalcopy 

\def\wacvPaperID{****} 

\ifwacvfinal\fi
\begin{document}

\title{Y-Autoencoders: disentangling latent representations via sequential-encoding}

\author{Massimiliano Patacchiola\thanks{Work done during an internship at Snap Inc.} \\
University of Edinburgh\\
{\tt\small mpatacch@ed.ac.uk}
\and
Patrick Fox-Roberts \\
Snap Inc.\\
{\tt\small pfoxroberts@snap.com}
\and
Edward Rosten \\
Snap Inc.\\
{\tt\small erosten@snap.com}
}

\maketitle
\ifwacvfinal\thispagestyle{empty}\fi

\begin{abstract}
In the last few years there have been important advancements in generative
models with the two dominant approaches being Generative Adversarial Networks
(GANs) and Variational Autoencoders (VAEs). However, standard Autoencoders
(AEs) and closely related structures have remained popular because they are
easy to train and adapt to different tasks. An interesting question is if we
can achieve state-of-the-art performance with AEs while retaining their good
properties. We propose an answer to this question by introducing a new model
called Y-Autoencoder (Y-AE). The structure and training procedure of a Y-AE
enclose a representation into an implicit and an explicit part. The implicit
part is similar to the output of an autoencoder and the explicit part is
strongly correlated with labels in the training set. The two parts are
separated in the latent space  by splitting the output of the encoder into two paths 
(forming a Y shape) before decoding and re-encoding. We then impose a number 
of losses, such as reconstruction loss, and a loss on dependence between the 
implicit and explicit parts. Additionally, the projection in the explicit 
manifold is monitored by a
predictor, that is embedded in the encoder and trained end-to-end with no
adversarial losses. We provide significant experimental results on various
domains, such as separation of style and content, image-to-image translation,
and inverse graphics.
\end{abstract}

\section{Introduction}


\begin{figure}
\centerline{\includegraphics[width=0.5\textwidth]{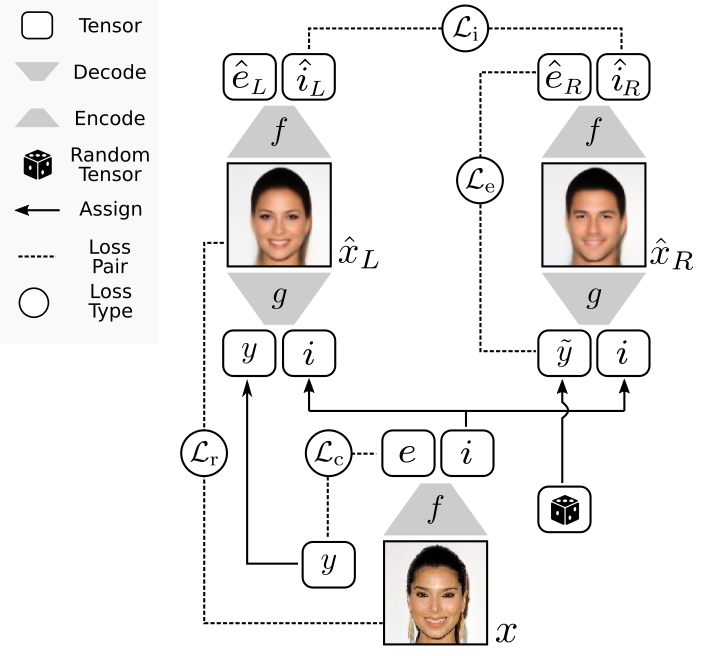}}
\caption{The structure and functioning of a Y-AE at training time. The
reconstruction is divided in two branches having shared parameters: left
(actual explicit content), and right (random explicit content). At test time
the two branches are merged resulting in a standard autoencoder.
}
\label{fig_training_testing}
\end{figure}

In this article we present a new training procedure for conditional autoencoders 
(cAE)  \cite{ballard1987modular, baldi1989neural} that allows a standard cAE
to obtain remarkable results in multiple conditional tasks. We call the
resulting model Y-Autoencoder (Y-AE), where the letter Y is a reference to the
particular branching structure used at training time. Y-AEs generally represent
explicit information via discrete latent units, and implicit information via
continuous units.

Consider the family of generative models where an input $x$ is conditioned 
on some paired explicit information $e$, such that we can encode the input
through the function $f(x)$ and then decode via $g(e, f(x))$ to generate new
samples $\hat{x}$. The explicit information can be any additional information 
about the inputs, such as labels, tags, or group assignments.  We may 
replace $f$ and $g$ by universal approximators, such as neural networks. 
Estimating the parameters of such a universal approximator describes the 
problem of fitting a cAE. 

While cAEs have met with some success, they often struggle disentangling the
latent representation. In other words the fitting procedure often ignores $e$,
since there is no effective regularization to enforce an effect.
For this reason, recent
work has mainly tackled conditional generation through Variational Autoencoders
(VAEs) \cite{kingma2013auto} and Generative Adversarial Networks (GANs)
\cite{goodfellow2014generative}. The former rely on a probabilistic approach,
that can capture relevant properties of the input space and constraint them
inside a latent distribution via variational inference. The latter are based on
a zero-sum learning rule, that simultaneously optimize a generator and a
discriminator until convergence. Both VAEs and GANs can be conditioned on the
explicit information $e$ to generate new samples $\hat{x}$. Recent work in this
direction has explored facial attributes generation \cite{choi2018stargan},
natural image descriptions \cite{dai2017towards}, people in clothing
\cite{lassner2017generative}, and video frame prediction \cite{liang2017dual}.

However, both VAEs and GANs suffer of a variety of problems. GANs 
are notoriously difficult to train, and may suffer of mode collapse when the state space 
is implicitly multimodal \cite{goodfellow2014generative}. VAEs rarely include discrete 
units due to the inability to apply backpropagation through those 
layers (since discrete sampling operations create discontinuities giving the objective
function zero gradient with respect to the weights), preventing the use of categorical
variables to represent discrete properties of the input space. Both have difficult 
exploiting rich prior problem structures, and must attempt to discover these 
structures by themselves, leaving valuable domain knowledge unused.

The Y-AE provides a flexible method to train standard cAEs that avoid these 
drawbacks, and can compete with specialized approaches. As there is no structural change the 
Y-AE simply becomes a cAE at test time, and it it is possible to assign values to 
the discrete units in the explicit layer whilst keeping the implicit
information unchanged. It is important to notice that for a Y-AE the definition
of implicit and explicit is very broad. The explicit information can either be
the label assigned to each element of the dataset, or a weak label that just
identifies a group assignment. 

The contribution of this article can be 
summarized in the following points:
\begin{enumerate}
\item Our core contribution is a new deep architecture called Y-AE, a
conditional generative model that can effectively disentangle implicit and
explicit information.
\item We define a new training procedure based on the sequential-encoding of
the reconstruction, which exploit weight sharing to enforce latent
disentanglement. This procedure is generic enough to be used in other contexts
or merged with other methods.
\item We perform quantitative and qualitative experiments to verify the
effectiveness of Y-AEs and the possibility of using them in a large variety of
domains with minimal adjustments.
\item We provide the open source code to reproduce the experiments. 
\footnote{\url{https://github.com/mpatacchiola/Y-AE}}
\end{enumerate}

\subsection{Previous work}




\textbf{Autoencoders.} Deep convolutional inverse graphics networks
(DC-IGNs) \cite{kulkarni2015deep} use a graphics code layer to disentangle
specific information (e.g. pose, light). DC-IGNs are bounded by the need
to organize the data into two sets of mini-batches, the first corresponding to
changes in only a single extrinsic variable, the second to changes in only the
intrinsic properties. In contrast Y-AEs do not require
such a rigid constraint.
Deforming Autoencoders \cite{shu2018deforming} disentangle shape from appearance through
the use of a deformable template. A spatial deformation warps the texture to 
the observed image coordinates.

\textbf{VAEs.} Conditional VAEs have been used in \cite{yan2016attribute2image} 
to produce a disentangled latent, allowing for model generation and 
attributes manipulation. A variant of VAEs, named beta-VAE \cite{higgins2017beta},
has showed state-of-the-art results in disentangled factor learning 
through the adoption of a hyperparameter $\beta$ that balances latent capacity 
and independence constraints. A cycle-consistent VAE was proposed in 
\cite{harsh2018disentangling}. This VAE is based on the minimization 
of the changes applied in a forward and reverse transform, 
given a pair of inputs.

\textbf{Adversarial.} Adversarial autoencoders (aversarial-AE)
\cite{makhzani2015adversarial} achieve disentanglement by matching the 
aggregated posterior of the hidden code vector 
with an arbitrary prior distribution and using a discriminator trained with an
adversarial loss. In \cite{hu2018disentangling} the disentangled representation is learned
through mixing and unmixing feature chunks in latent space with the supervision 
of an adversarial loss. The authors also use a form of sequential encoding 
that has some similarities with the one we propose. However, the key difference 
is that in a Y-AE part of the latent information is explicit
and controllable whereas in \cite{hu2018disentangling} is not.

\textbf{GANs.} A conditional form of GAN has been
introduced in \cite{mirza2014conditional}, and it is constructed such that
the explicit information is passed to both the generator and discriminator.
\cite{chen2016infogan} propose 
InfoGAN, an information-theoretic extension of GANs, 
with the aim of performing unsupervised disentanglement.
\cite{zhu2017unpaired} used a type of GAN named CycleGAN to 
concurrently optimize two mapping functions and two discriminators through an
adversarial loss. Differently from CycleGANs, Y-AEs rely on a single network,
and the consistency is ensured in the latent space with the aim of maximizing
the distance between domains in the image space.



\section{Description of the method}\label{sec_description_method}

\subsection{Notation}\label{ssec_notation}
We define an autoencoder as a neural network consisting of two parts: encoder
and decoder. The encoder is a function $f$ performing a non-linear mapping
$\mathbb{R}^{n} \mapsto \mathbb{R}^{m}$ from an input $x\in\mathbb{R}^n$ to a latent
representation $h\in\mathbb{R}^m$. We refer to this latent representation as the code. The
decoder is a function $g$ performing a non-linear mapping $\mathbb{R}^{m}
\mapsto \mathbb{R}^{n}$ from the latent representation $h$ to a reconstruction
$\hat{x}$. Encoder $f_\theta(x)$ and decoder $g_\psi(h)$ are parametrized by
$\theta$ and $\psi$ respectively - these are omitted in the rest of the article
to keep the notation uncluttered. Parameters are adjusted during an online
training phase via stochastic gradient descent, minimizing the mean squared
error between the input and reconstruction $\mathcal{L} = \rvert g(f(x)) - x \rvert^{2}$ on
a random mini-batch of samples. The network is designed such that $n \gg m$,
forming a bottleneck. This ensures that if $\mathcal{L}$ is small, $h$ must be
a compressed version of $x$.

This article focuses on the particular case where we have access to some label
information $y$, and have divided the latent representations $h$ into two parts
$h=\{i,e\}$, where the $i$ stands for implicit and $e$ stands for explicit. The
distinction between the two is that the explicit information should be
approximately equal to $y$, whereas the implicit information should be
independent of it. We denote a decoder which takes a separable hidden state as
input by $g(i, e)$. The label $y$ may take the form of a one-hot vector (in a
classification setting) or a vector of real values (in a regression).

\subsection{Overview}
The encoding phase of a Y-AE is identical to a standard cAE, but the
reconstruction is quite different as it is divided into two branches, left and
right. These two branches share the same weights, similarly to a siamese
network \cite{bromley1994signature}. Thus, the Y-AE requires no more parameters
than its cAE counterpart (see Figure \ref{fig_training_testing}).

The implicit information $i$ produced by $f(x)$ is given as input to the two
branches, whereas the explicit information $e$ is discarded. Instead of $e$ the
left branch takes as input the actual label $y$ and the right branch takes as
input a random label $\tilde{y}$. The decoding phase produces two
reconstructions $\hat{x}_{L}$ and $\hat{x}_{R}$, where the subscript $L$
specifies the left branch, and the subscript $R$ the right branch. At this point
we have two images which have identical implicit representations but different
explicit ones.

\begin{figure}
\centerline{\includegraphics[width=\columnwidth]{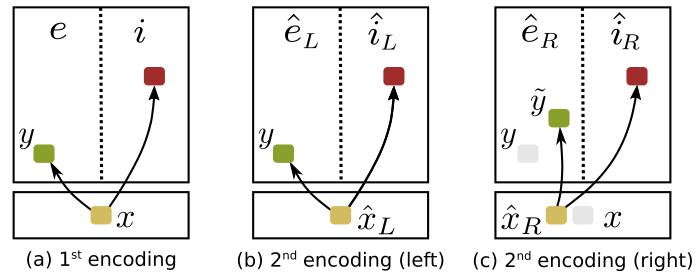}}
\caption{Graphical representation of sequential-encoding. (a) In the first
encoding phase the input vector (yellow square) is encoded in two manifolds:
explicit (green square) and implicit (red square). In the second encoding
phase, the two branches have different purposes. (b) In the left branch the
content is not manipulated and the latent representation has to be consistent
with the first phase. (c) In the right branch the explicit content is
manipulated meaning that it should change in accordance with the manipulation,
while the implicit code should stay the same.}
\label{fig_latent_space_sequential_encoding}
\end{figure}

The encoding stage is then applied to the two branches (a process we call
sequential-encoding), producing two latent representations $\{\hat{i}_{L}$,
$\hat{e}_{L}\}$, and $\{\hat{i}_{R}$, $\hat{e}_{R}\}$. The sequential encoding
is used to verify that the implicit representations are not altered when only
the explicit repesentation changes. In addition, the right branch ensures that
the explicit information is not also hidden in the implicit data, since it must
be able to propagate through (see Figure
\ref{fig_latent_space_sequential_encoding}). It is important to notice that the
sequential-encoding is only applied at training time as showed in Figure
\ref{fig_training_testing}. The second encoding stage concludes the forward
pass. The backward pass is based on the simultaneous optimization of multiple
loss function and it is described in the next section.

\subsection{Definition of the loss function}

The loss function consists of four separate components.

Firstly, in the left branch of a Y-AE the label $y$ is assigned explicitly,
replacing the $e$ component inferred by the encoder. This is done to avoid
instability in the preliminary learning phase, when the classifier
predictions are still inaccurate. To ensure appropriate reconstructions
following this we penalize deviations between $\hat{x}_{L}$ and $x$, using the
standard least-squared error reconstruction loss:

\begin{equation}\label{eq_left_reconstruction_loss}
\mathcal{L}_\text{r} = \rvert\hat{x}_{L} - x\rvert^{2},
\end{equation}
where $\hat{x}_{L} = g(i, y)$.

Secondly, we include a computationally cheap cross-entropy loss penalty between $y$ and $e$,
\begin{equation}\label{eq_central_predictor_loss}
\mathcal{L}_c = CE(e, y) = -\sum_j e_j \log y_j.
\end{equation}
This is done as this particular part of the output of the encoder can be
considered as the output of a predictor, identifying which type of explicit
content is present in the input $x$. 

This predictor-like aspect is made direct use of in the right branch of the
Y-AE, where it is deployed to verify the consistency of the relation
$\hat{e}_{R} \approx \tilde{y}$. This is ensured using a third, cross-entropy
loss:

\begin{equation}\label{eq_left_explicit_loss}
\mathcal{L}_e = CE(\hat{e}_{R}, \tilde{y}) = -\sum_j \hat{e}_{R,j} \log \tilde{y}_j.
\end{equation}

Finally, on the left branch a sequential-encoding is also performed. The vector
$\hat{i}_{L}$ can be compared with $\hat{i}_{R}$ the right counterpart. Since
the implicit information has not been manipulated it should be consistent in
the two branches. This constraint can be added as an Euclidean distance
penalty:

\begin{equation}\label{eq_euclidean_loss}
\mathcal{L}_i = d(\hat{i}_{L}, \hat{i}_{R}) = \lVert \hat{i}_{L} - \hat{i}_{R} \rVert_{2}.
\end{equation}

The losses defined above are then integrated in the global loss function

\begin{equation}\label{eq_global_loss}
\mathcal{L} = \mathcal{L}_\text{r}
            + \mathcal{L}_c\\
            + \lambda_e \mathcal{L}_e\\
            + \lambda_i \mathcal{L}_i\\,
\end{equation}
where the relative contribution of the explicit and implicit losses can be
controlled by altering $\lambda_e$ and $\lambda_i$ respectively. Note that the
reconstruction and classification losses have not been given similar
weightings, since the first is the main reconstruction objective, and the
second only acts in support of the explicit loss (which is already accounted
for). An ablation study of the effect of altering $\lambda_e$ and $\lambda_i$
is presented in the experimental section (Section \ref{ssec_ablation_study}).

\section{Experiments}\label{sec_experiments}

In order to demonstrate the efficacy of the Y-AE training scheme, we use a
straightforward autoencoder architecture.  In Section~\ref{ssec_ablation_study},
we remove various parts of the Y-AE structure to show that they are all
necessary, in Section~\ref{ssec_comparison_against_baselines}, we compare the
Y-AE training method to some simpler baseline training methods and in
Section~\ref{ssec_cross_domain_evaluation}, we evaluate the Y-AE structure on a
variety of different tasks in a qualitative manner to show it's applicability to
a variety of domains.

\subsection{Implementation}

The encoders used in these experiments are based on the principle of
simultaneously halving the spatial domain whilst doubling the channel domain,
as successfully used by \cite{kulkarni2015deep} (the opposite is done in the
decoding phase). Each network module is made of three consecutive operations:
convolution (or transpose convolution in the decoder), batch normalization, and
leaky-ReLU. For inputs of size $32 \times 32$ four such modules have been used,
increasing to six for inputs of size $128 \times 128$. Reduction (or
augmentation) is performed via stride-2 convolution (or transpose convolution).
No pooling operations have been used at any stage. The input images have been
normalized so to have continuous values $\in \big[0, 1 \big]$.
The sigmoid activation function has been used in the implicit portion of the
code, and softmax in the explicit part. All the other units use a leaky-ReLU
with slope of $0.2$. The parameters have been initialized following the Glorot
uniform initialization scheme \cite{glorot2010understanding}.  To stabilize the
training in the first iterations, we initialized the parameters of the input to
the implicit layer's sigmoid activation function by randomly sampling from a
Gaussian distribution ($\mu=0$, $\sigma^{2}=0.01$) with the bias set to a
negative value ($-5$) such that the sigmoid is initially saturated toward zero.
All the models have been trained using the Adam optimizer
\cite{kingma2014adam}. The models have been implemented in Python using the
Tensorflow library, and trained using a cluster of NVIDIA GPUs of the following
families: TITAN-X, K-80, and GTX-1060. A detailed description of the networks
structure and hyperparameters is reported in the supplementary material.

\begin{figure*}
\centerline{\includegraphics[width=\textwidth]{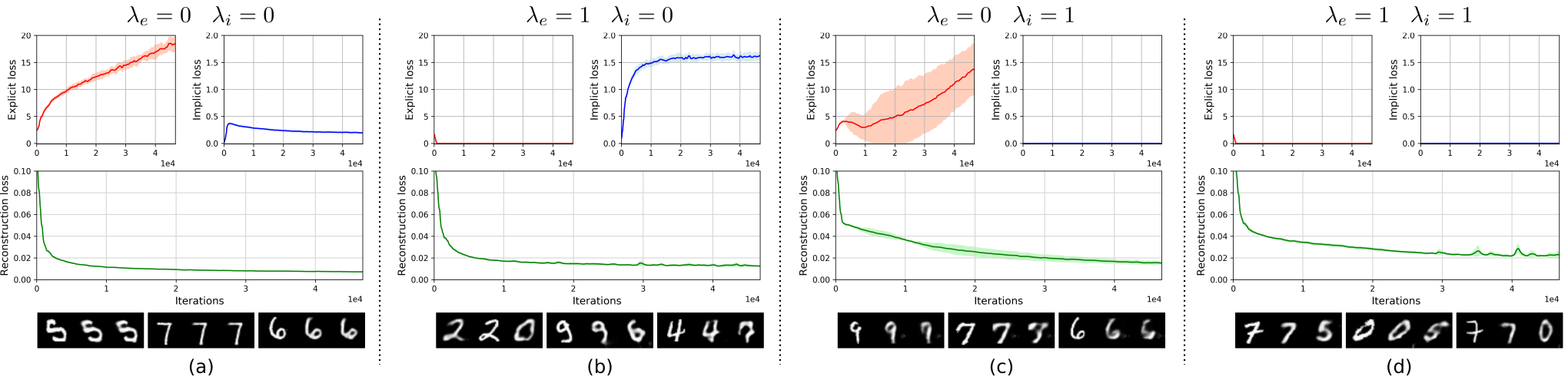}}
\caption{Ablation study for a Y-AE trained on the MNIST
dataset. Top: 
plots of
the explicit (red), implicit (blue) and reconstruction (green) loss for $4.6
\times 10^{4}$ iterations (100 epochs, 3 runs). Note that the plots have the
same scale. Bottom: three random samples from the last iteration, for
each group we report the input (left digit), reconstruction (central digit),
and reconstruction with random content (right digit).
(a) Explicit and implicit loss are not minimized; note that the digit content
cannot be changed.
(b) Only the
explicit loss is minimized; changing the content alters the style
(c) Only the implicit loss is minimized; changing the content works very poorly (d) Both
explicit and implicit losses are minimized; the content can be changed without
altering the style.
\label{fig_experiments_ablation}}
\end{figure*}

\subsection{Ablation study} \label{ssec_ablation_study}

In this experimental section we compare against different ablations of the full
loss (Equation \ref{eq_euclidean_loss}), to provide a deeper understanding of
the results presented in Section \ref{ssec_cross_domain_evaluation}. To do this
we vary the mixing coefficients $\lambda_e$ and $\lambda_i$ that regulate the
weight of the explicit (Equation \ref{eq_left_explicit_loss}) and the implicit
(Equation \ref{eq_euclidean_loss}) losses, systematically setting each to $0$
or $1$. As such, we either minimize neither of the losses ($\lambda_e=0$,
$\lambda_i=0$), only the explicit loss  ($\lambda_e=1$, $\lambda_i=0$), only
the implicit loss  ($\lambda_e=0$, $\lambda_i=1$), or both ($\lambda_e=1$,
$\lambda_i=1$). 

The focus of this section is a widely used benchmark: the Modified National
Institute of Standards and Technology (MNIST) \cite{lecun1998gradient}. This
dataset is composed of a training set of $60000$ greyscale handwritten digit
images and a test set of $10000$ images. We aim at separating an implicit
information such as the style of the digits (orientation, stroke, size, etc)
and an explicit information represented by the digit value (from 0 to 9). We
used a Y-AE with 32 units in the implicit portion of the code and 10 unit in
the explicit portion. We trained the model for 100 epochs using the Adam
optimizer with a learning rate of $10^{-4}$ and no weight decay. The results
are the average of three runs for each condition.


\begin{table}
\begin{center}
\begin{tabular}{ccccc}
\cline{2-5}
& \begin{tabular}{@{}c@{}}$\lambda_e=0$ \\ $\lambda_i=0$\end{tabular} & \begin{tabular}{@{}c@{}}$\lambda_e=1$ \\ $\lambda_i=0$\end{tabular} & \begin{tabular}{@{}c@{}}$\lambda_e=0$ \\ $\lambda_i=1$\end{tabular} & \begin{tabular}{@{}c@{}}$\lambda_e=1$ \\ $\lambda_i=1$\end{tabular}\\
\hline
$\mathcal{L}$     & $0.066$  & $0.075$            & $0.081$            & $0.089$\\
$\mathcal{L}_{r}$ & $0.007$  & $0.012$            & $0.015$            & $0.023$\\
$\mathcal{L}_{c}$ & $0.059$  & $0.063$            & $0.066$            & $0.064$\\
$\mathcal{L}_{e}$ & $18.9$   & $3 \times 10^{-4}$ & $14.0$             & $8 \times 10^{-5}$\\
$\mathcal{L}_{i}$ & $0.197$  & $1.58$             & $6 \times 10^{-4}$ & $0.002$\\
\hline
\end{tabular}
\end{center}
\caption{Comparison of the average loss value (three runs) on the 10000 images
of the MNIST test set for all the ablation conditions. $\mathcal{L}$ is the
global loss. $\mathcal{L}_{r}$ is the reconstruction loss (MSE between $x$ and
$\hat{x}_{L}$). $\mathcal{L}_{c}$ is the classification loss. $\mathcal{L}_{e}$
is the explicit loss. $\mathcal{L}_{i}$ is the implicit loss.}
\label{tab_ablation_test_results}
\end{table}

An overview of the results is reported in Figure \ref{fig_experiments_ablation}
and the average loss on the test set in Table \ref{tab_ablation_test_results}.
In the first condition ($\lambda_e=0$, $\lambda_i=0$; Figure
\ref{fig_experiments_ablation}-a) there are no constraints on the two losses
and the reconstruction reach a value of $0.007$ on the test set. This
is achieved by exploiting the implicit portion of the code and ignoring the
information carried by the explicit part (note the large disparity in implicit
and explicit loss). The triplet of samples in the first column show
that the explicit information has been ignored as each displays three almost
identical digits, indicating the two branches of the Y-AE produce identical
outputs. In the second condition ($\lambda_e=1$, $\lambda_i=0$; Figure
\ref{fig_experiments_ablation}-b) only the explicit loss is regularized. This
forces the output of the right branch to take account of the explicit
information. However, as there is no regularization on the implicit portion of
the code, the network learns to use it to carry the explicit information. The
samples produced by the right branch show that the explicit content has been
kept but the style partially corrupted. This condition has been further
investigated in Section \ref{ssec_comparison_against_baselines}. The third case
($\lambda_e=0$, $\lambda_i=1$; Figure \ref{fig_experiments_ablation}-c) only
regularizes the implicit loss. The explicit loss rapidly diverges, indicating
that the reconstruction on the right branch does not resemble the digit it
ought to. The samples confirm this assumption, showing the right
reconstructions as more similar to the inputs than to the random content.
Finally, the fourth and last condition ($\lambda_e=1$, $\lambda_i=1$; Figure
\ref{fig_experiments_ablation}-d) is the complete loss function with all the
components being minimized. Both explicit and implicit losses rapidly converge
toward zero, whereas the reconstruction loss moves down reaching a value of
$0.023$ on the test set. The samples produced clearly show that the style of
the input (left digit) is kept and the content changed (right digit).

An overall comparison between all the conditions shows that the implicit loss
(blue curve) act as a regularizer, with the effect of inhibiting the
reconstruction on the left branch. This is an expected result, since the
implicit loss limits the capacity of the code and it ensures that only the
high-level information about the style is considered. A qualitative analysis of
the samples shows that only the use of both explicit and implicit losses
(Figure \ref{fig_experiments_ablation}-d) guarantees the disentanglement of
style and content, supporting our hypothesis about the functioning of the Y-AE.
In particular it is evident how the high-level style information has been
correctly codified, with the generated samples incorporating orientation,
stroke, and size of the inputs.

\subsection{Comparison against baselines} \label{ssec_comparison_against_baselines}

In this section we compare the proposed method against different baselines on
the MNIST dataset. In all cases, we encode the input image, change the explicit
information (i.e. the number), then decode it to produce an image. We use a
a pre-trained classifier to test whether the generated images have the right
appearance. Also, since changing $e$ should change the digits, 
we test similarity of the generated digits against the
original digits using MSE (which should be high) and the perceptual structural
similarity measure, SSIM\cite{wang2004image}, which should be low.

We train the autoencoder 
models using the same set-up described in
Section \ref{ssec_ablation_study}. The evaluation has been performed encoding
all the inputs $x$ in the test set, extracting the implicit code $i$, and
randomly sampling (without replacement) $60\%$ of the possible contents
$\widetilde{y}$, then $i$ and $\widetilde{y}$ were used to get the
reconstruction $\hat{x}$. This procedure generated a dataset $D=\{ t_{1}, ...,
t_{N} \}$ with $t$ being input-label tuples $t=(\hat{x}, \widetilde{y})$, and
$N$ being six times larger than the original test set. For the evaluation classifier, 
we train an ensemble of five LeNet \cite{lecun1998gradient} classifiers on the original
dataset.


%
%
%
%

\begin{table}
\begin{center}
\begin{tabular}{lccc}
\hline
\textbf{Method} & \textbf{Accuracy} (\%) & \textbf{SSIM} & \textbf{MSE}\\
\hline
cAE & $10.6 \pm 0.1$  & $0.87$ & $17.52$\\
cAE  + regularizer & $66.9 \pm 17.5$ & $0.55$ & $26.43$\\
adversarial-AE \cite{makhzani2015adversarial} & $43.4 \pm 10.5$ & $0.57$ & $27.4$\\
cVAE \cite{kingma2013auto} & $96.7 \pm 1.6$ & $0.50$ & $27.05$\\
beta-VAE \cite{higgins2017beta} & $\boldsymbol{99.7 \pm 0.1}$ & $0.42$ & $30.43$\\
Y-AE + ablation [our] & $90.5 \pm 2.9$  & $0.59$ & $27.38$\\
Y-AE [our] & $99.5 \pm 0.1$  & $\boldsymbol{0.37}$ & $\boldsymbol{42.99}$\\
\hline
\end{tabular}
\end{center}
\caption{Comparison of different methods for multiple metrics. Accuracy of the
explicit reconstruction (percentage), measured through an independent ensemble
of classifiers trained on the MNIST dataset. Internal SSIM and MSE between the
input sample $x$ and all the 10 reconstructions $\hat{x}$. Notice that optimal
internal metrics should have low SSIM and high MSE meaning that the samples are
different from the input. The results are the average of three runs.
Best results highlighted in bold.}
\label{tab_comparison_test_results}
\end{table}

\begin{figure}
\centerline{\includegraphics[width=\columnwidth]{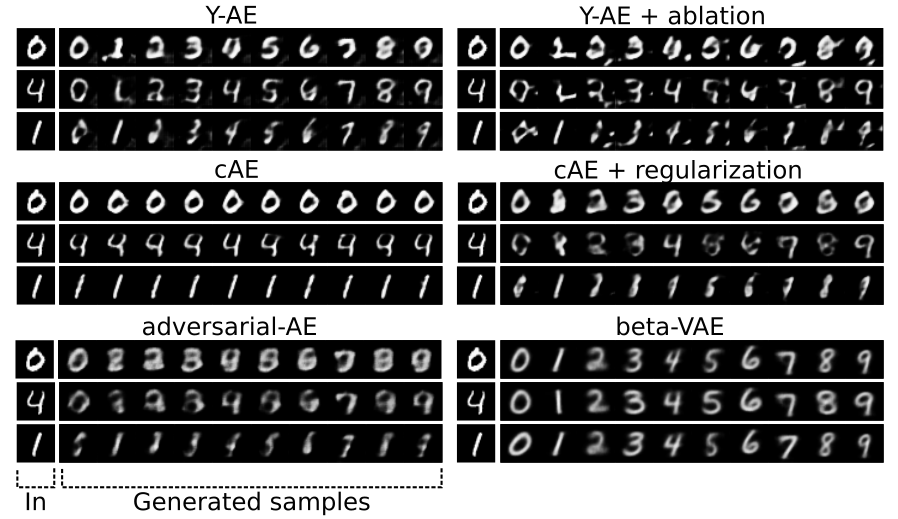}}
\caption{Samples produced at test time by different methods given the same
inputs (leftmost column). Not cherry picked. The Y-AE is able to use both 
implicit (style) and explicit (digit value) information. 
As expected, the Y-AE performance is affected by the removal of the implicit loss, 
showing the importance of both losses.
The cAE and the adversarial-AE discard the explicit information loosing 
the ability of changing the content. The VAE has a poor style transfer, 
especially when the implicit information is pushed closer to the Gaussian prior through $\beta$.}
\label{fig_comparison_samples}
\end{figure}

\textbf{Conditional Autoencoders (cAEs).} A cAE defines a conditional
dependence on explicit information $e$ such that the reconstruction $\hat{x}$
is conditioned on both the input $x$ and the labels $y$. This can be considered
as the main baseline, since a cAE has the same structure of a Y-AE but only
relies on a standard MSE loss (see Equation \ref{eq_left_reconstruction_loss})
to minimize the distance between the inputs and the reconstructions. 

\textbf{cAE + regularization.} We applied a series of regularizers on cAEs to
push their performance. Strong regularization may enforce disentanglement of
style and content limiting the amount of information codified in the latent
space. We drastically reduced the number of epochs from 100 to
20, the number of implicit units from 32 to 16, and we applied a weight decay
of $10^{-4}$. None of these regularizers has been used in the Y-AE.

\textbf{Adversarial-AE.} We performed a comparison against an adversarial-AE 
\cite{makhzani2015adversarial}. As adversarial discriminator we used a multi-layer 
perceptron with 512 hidden units and leaky-ReLU. It was necessary 
to apply strong regularization in order to obtain decent results 
(8 units as code, 20 epochs, $10^{-4}$ weight decay).

\textbf{VAE.} We also compared the method against a 
conditional VAE (cVAE) \cite{kingma2013auto} and a 
beta-VAE \cite{higgins2017beta} ($\beta=2$).

\textbf{Y-AE + ablation.} To check whether the Y-AE accuracy is just a result
of the fact that it has been trained with a predictor in the loop, we
tested against the ablated version of the model with $\lambda_e=1$ and
$\lambda_i=0$. This condition produces samples with consistent content but the
style can be partially corrupted (Section \ref{ssec_ablation_study}). We expect
to see the accuracy being lower than the Y-AE
trained with the complete loss function, because in comparison the samples have
lower quality.

The quantitative results are reported in Table
\ref{tab_comparison_test_results}, and the qualitative results in Figure
\ref{fig_comparison_samples}.
The accuracy of the Y-AEs is higher than most of the other methods ($99.5 \%$),
meaning that the samples carry the right content. As a result we observe that
the SSIM is low ($0.37$) and the MSE high ($42.99$), as expected when style and
content are well separated. Conversely, the accuracy of standard cAEs is
close to chance level ($10.6 \%$) because the model is 
producing the same digit (Figure \ref{fig_comparison_samples}-cAE) 
and ignoring the content information $e$, 
meaning that just $10\%$ of the produced samples are correct. 
Interestingly, the accuracy of Y-AEs with ablation is pretty high
($90.5 \%$) but inferior to the standard counterpart, 
with the samples showing stylistic artifacts caused by the ablation of the implicit loss.
Strong regularization 
increases the performances of cAEs but the results are
still far from both standard and ablated Y-AEs. 
The performance of the cVAE and beta-VAE is lower in SSIM 
and MSE when compared to the Y-AE, with beta-VAE being slightly better in terms 
of accuracy ($99.7\%$ vs $99.5\%$).
The digits produced by the beta-VAE are clear but the style does not significantly 
change among the inputs (Figure \ref{fig_comparison_samples}-beta-VAE). This is due to 
the pressure imposed by $\beta$ on the Kullback-Leibler divergence 
that moves the latent space closer to the Gaussian prior resulting in 
low expressivity. In conclusion, the qualitative analysis of the
samples (Figure \ref{fig_comparison_samples}) shows that Y-AEs are
superior to other methods on the problem at hand (see Section
\ref{ssec_cross_domain_evaluation} for additional samples).

\subsection{Cross-domain evaluation} \label{ssec_cross_domain_evaluation}

The evaluation of the method has been done in three ways in order to
verify the performances on a wide set of problems. The first test is
disentanglement of style and content, the second pose generation (inverse graphics), 
and the third unpaired image-to-image translation.

\begin{figure}
\centerline{\includegraphics[width=\columnwidth]{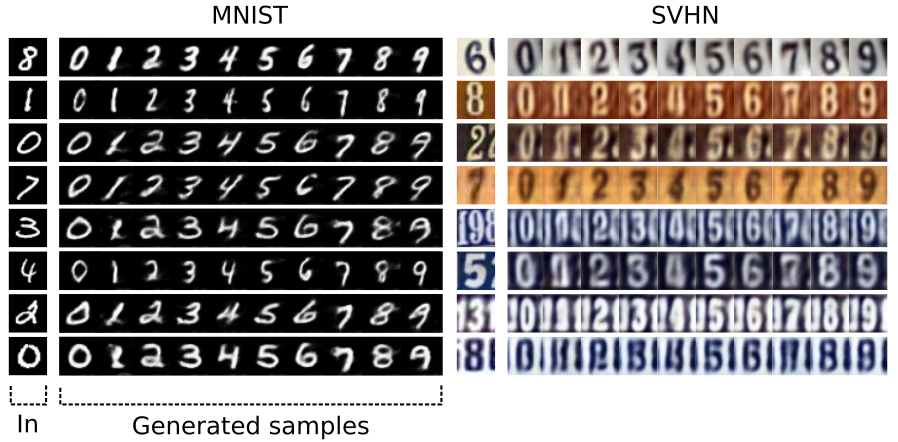}}
\caption{Samples produced at test time by a Y-AE trained on the MNIST and SVHN
datasets. The first column is the input, other columns are the reconstructions
given all possible content values. The Y-AE is able to keep the style
(orientation, stroke, height, width, etc) and change the content.}
\label{fig_experiments_mnist_svhn}
\end{figure}

\textbf{Disentanglement of style and content.} This experiment shows how a Y-AE 
can be used to disentangle style and content. This is
shown through two widely used datasets: the MNIST \cite{lecun1998gradient},
and the Street View House Number (SVHN) \cite{netzer2011reading}. In this task,
the implicit information is the style (orientation, height,
width, etc) and the explicit information is the content (digit value).
We set $\lambda_{e}=1$ and
$\lambda_{i}=0.5$ on the MNIST and $\lambda_{e}=1$ and $\lambda_{i}=0.1$ on the
SVHN dataset. We report some of the generated samples in Figure
\ref{fig_experiments_mnist_svhn}. In the MNIST the implicit units have captured
the most important underlying properties, such as orientation, size, and
stroke. Similarly on the SVHN dataset the model has been able to retain the
explicit information and codify the salient properties (digit style, background
and foreground colours) in the implicit portion of the code.

\begin{figure*}
\centerline{\includegraphics[width=\textwidth]{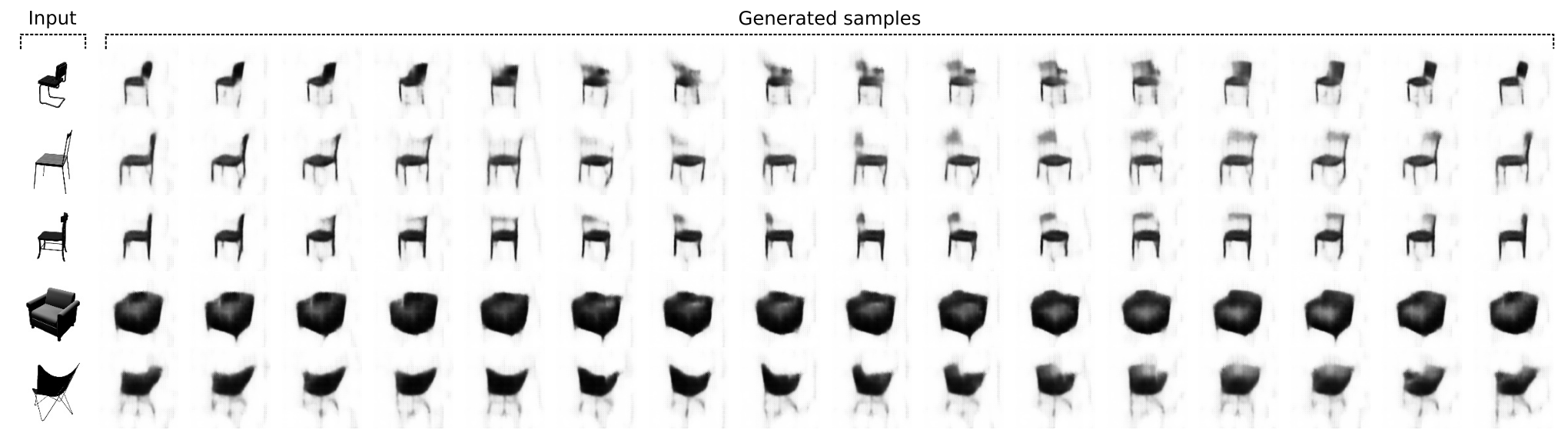}}
\caption{Samples produced by a Y-AE trained on the chairs dataset. Given the
implicit encoding obtained through an input chair type (leftmost column) it is
possible to generate a rotation of 360 degrees activating the corresponding
discrete units in the explicit portion of the code and then decode. The network
has never seen these chairs before at any orientation. For graphical
constraints each row only shows 16 of the 31 generated poses.}
\label{fig_experiments_chairs}
\end{figure*}

\begin{figure}
\centerline{\includegraphics[width=\columnwidth]{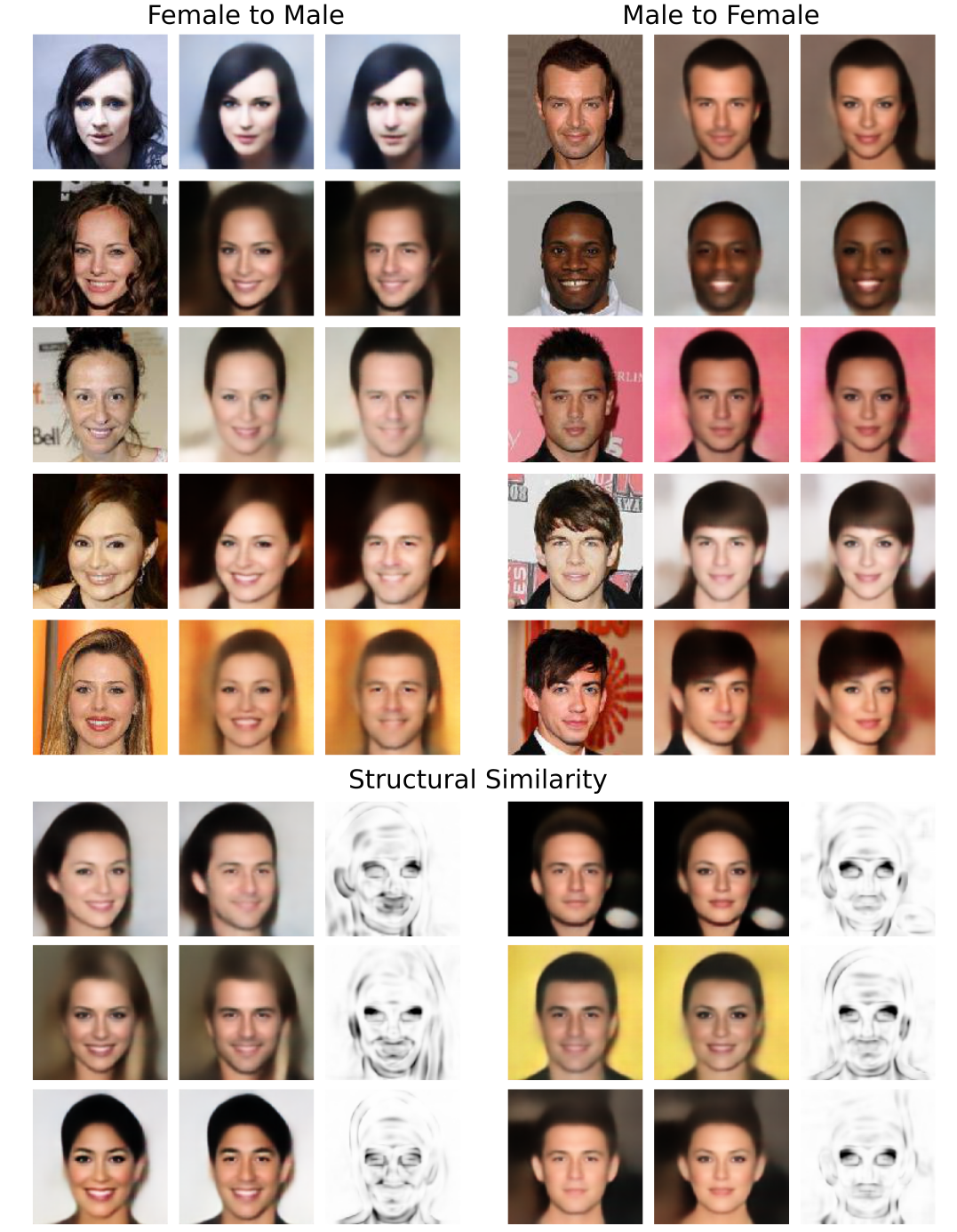}}
\caption{Unpaired image-to-image translation (male$\leftrightarrow$female). Samples produced by
a Y-AE trained on unpaired images of celebrities. Top: given the input image
(left column) it is possible to produce a reconstruction (central column) or to
change the sex (right column) manipulating the two units in the explicit
portion of the code. Bottom: given a reconstruction (left column) and the
reconstruction with switched sex (central column), we estimated the mean
structural similarity between the two (right column) and highlight in gray the
areas with major changes.}
\label{fig_experiments_male_female}
\end{figure}

\textbf{Pose generation (inverse graphics).} Pose generation consists in
producing a complete sequence of poses given a single frame of the sequence.
This task is particularly challenging because relevant details of the object
may be occluded in the input frame, and the network has to make a conjecture
about any missing component. We tested the Y-AE on the 3D chairs dataset
\cite{aubry2014seeing}. This dataset contains 1393 rendered models of chairs.
Each model has two sequences of 31 frames representing a 360 degrees rotation
around the vertical axis. Following a similar procedure reported in
\cite{kulkarni2015deep} we randomly selected 100 models and we used them as
test set. In a similar way we also preprocessed the images, first we removed 50
pixels from each border, then we resized the images to $128 \times 128$ pixels
greyscale using a bicubic interpolation over $4 \times 4$ pixel neighborhood.
The explicit representation has been encoded in the Y-AE using 31 units, one
unit for each discrete pose. The implicit information has been encoded with 481
units and used to codify the properties of the chair model. Results are showed
in Figure \ref{fig_experiments_chairs}. Even though the network never seen the
test model before, at any orientation, it is able to generalize effectively and
produce a full rotation of 360 degrees.

\begin{figure}
\centerline{\includegraphics[width=\columnwidth]{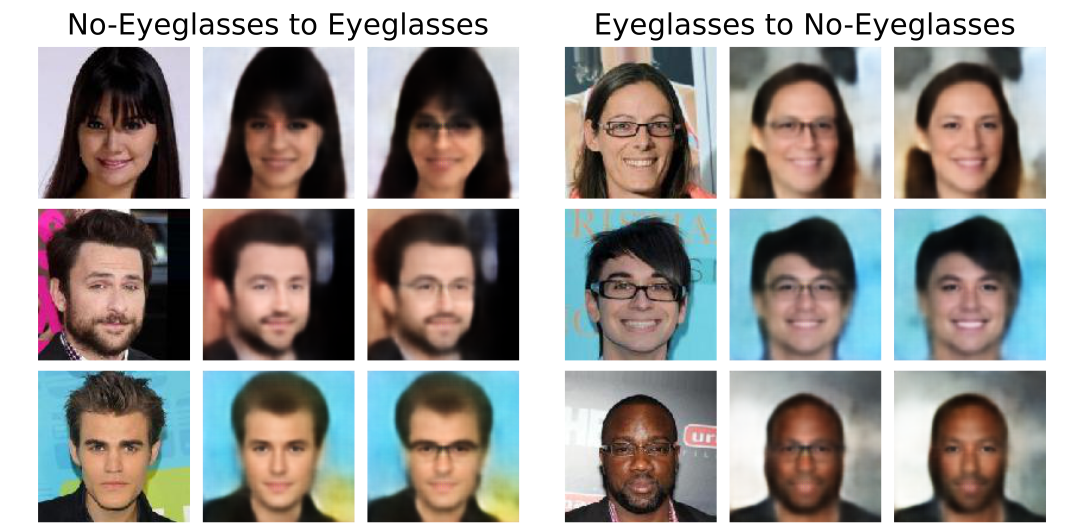}}
\caption{Samples produced by a Y-AE trained on image-to-image translation of a
facial attribute (eyeglasses). Each group of images is divided in three
columns: the left is the original image, the central is the reconstruction, and
the right is the reconstruction with different attribute. Note that the
reconstructions are more blurred compared to other samples because of the use
of an $L_{2}$ norm.}
\label{fig_experiments_facial_attributes}
\end{figure}

\textbf{Unpaired image-to-image translation.} The aim of this series of
experiments is to verify how Y-AEs behave when the explicit information is only
provided by a weak label, being the group assignment. In unpaired training
there are two separate sets $A$ and $B$ of input and targets that do not
overlap, meaning that samples belonging to $A$ are not present in $B$ and
vice-versa. The goal is to translate an image from one set to the other. Here
we focus on three particular types of unpaired translation problems:
male$\leftrightarrow$female,  glasses$\leftrightarrow$no-glasses, and
segmented$\leftrightarrow$natural. For the male$\leftrightarrow$female and
glasses$\leftrightarrow$no-glasses tasks we used the CelebA dataset
\cite{liu2015faceattributes}, a database with more than 200K celebrity images,
of size $128 \times 128$ pixels, each with 40 attribute annotations. 
The images cover large pose variations and
have rich annotations (gender, eyeglasses, moustache, face shape, etc). In the
male$\leftrightarrow$female task we used an $L_{1}$ penalty on the
reconstruction which generally gives sharper results. In the
glasses$\leftrightarrow$no-glasses we used instead an $L_{2}$ reconstruction
loss, so to compare the quality of the samples with both losses. We used
a neural networks with $1022$ implicit units and $2$ explicit units. 
For the problem of segmented$\leftrightarrow$natural translation we used a subset 
of the Cityscapes dataset \cite{cordts2016cityscapes}. Cityscapes is based on a
stereo video sequences recorded in streets from 50 different
cities, and it includes both natural and semantically segmented images. To have
unpaired samples, we randomly removed from the dataset one of the pair, so to
have half natural and half segmented images. The final dataset consisted of $2975$ 
training images of size $128 \times 128$ pixels. This is a fairly limited amount of images,
but we considered it as an additional test to verify the performance of the method 
on a limited amount of data. As regularization we just reduced the size of the 
code from $1022$ to $510$ implicit units.

\begin{figure}
\centerline{\includegraphics[width=\columnwidth]{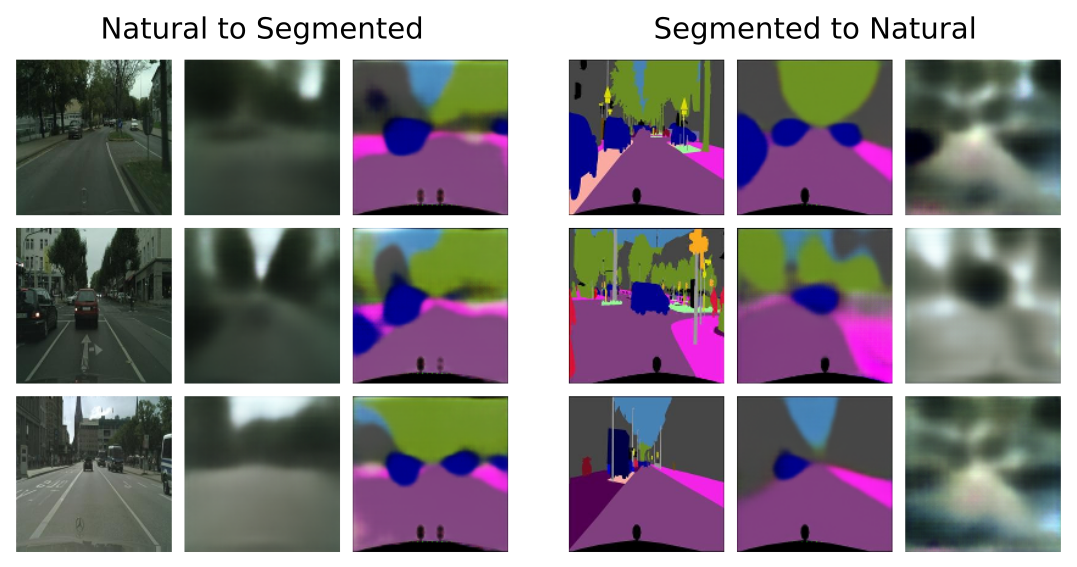}}
\caption{Unpaired image-to-image translation (segmented$\leftrightarrow$natural). Samples
produced by a Y-AE trained on unpaired images of segmented and natural
cityscapes. Given the input image (left column) it is possible to produce a
reconstruction (central column) or to change the reconstruction type (right
column) by manipulating the two explicit units in the code.
The passage from one domain to the
other is not always consistent, possibly indicating overfitting problems.} 
\label{fig_experiments_segmentation_natural}
\end{figure}

Results obtained on the CelebA dataset are reported in Figure
\ref{fig_experiments_male_female} and Figure
\ref{fig_experiments_facial_attributes}. In the male$\leftrightarrow$female
task, the transition from one sex to the other looks robust in most of the
samples (Figure \ref{fig_experiments_male_female}-top). To understand which
attributes have been codified in the implicit portion of the code we computed
the SSIM between the two reconstructions. We report in Figure
\ref{fig_experiments_male_female}-bottom the greyscale maps based on this
metric. The SSIM shows that the most intense changes are localized in the eyes
region, with a peak on eyebrows and eyelashes. Minor adjustments are applied
around the mouth (lips and beard), forefront (wrinkles), cheekbones, ears, and
hairline. It is important to notice that the model found these differences
without any specific supervision, only through the weak labelling identifying
the groups.
In the glasses$\leftrightarrow$no-glasses task, the samples look more blurred
because of the use of the $L_{2}$ loss, however also in this case the
transition is robust.
The Cityscapes dataset (Figure \ref{fig_experiments_segmentation_natural}) 
translation task proved extremely challenging. 
The model has been able to successfully separate the two domains and 
identify the major factors of variation (e.g. sky, road, cars, trees, etc). 
However, minor details such as road signs and vehicle type have been discarded.
We suspect this is due to the 
large difference between the two domains, small number of images, 
and highly lossy nature of the natural$\rightarrow$segmented translation.  
Further work is required to overcome the difficulties presented in this setting.

\section{Discussion and conclusions}

In this article we present a new deep model called Y-AE, allowing
disentanglement of implicit and explicit information in the latent space without
using variational methods or adversarial losses. The method splits the
reconstruction in two branches (with shared weights) and performs a
sequential encoding, with an implicit and an explicit loss ensuring the
consistency of the representations. We show through a wide experimental
session that the method is effective and that its performance is superior to
similar methods.



Future work should mainly focus on applying the principles of Y-AEs to GANs and
VAEs. For instance, codifying the implicit information as a Gaussian distribution it is
possible to integrate Y-AEs and VAEs in a unified framework and having the best of both worlds.


{\small
\bibliographystyle{ieee}

}



\section*{Supplementary material}

# Appendices

## A. Network architectures

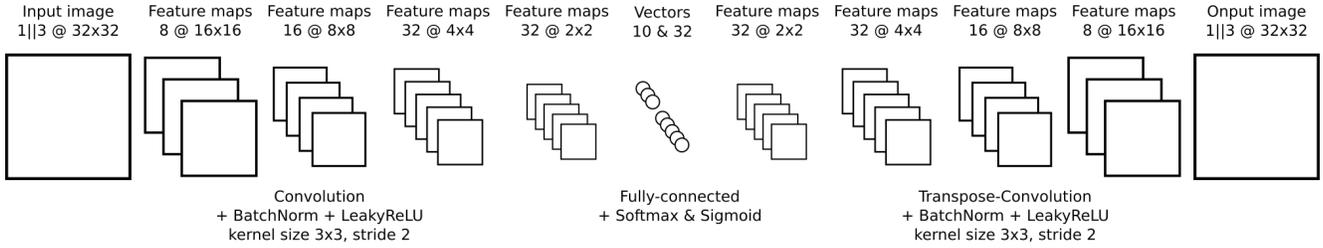

Figure 1: Neural network structure used in the MNIST and SVHN experiments. Notice that to avoid clutter just some of the feature maps have been reported. The notation represents: channels @ feature maps (or vectors) size. Vertical bars || represent logical OR disjunction for the two datasets (e.g. 1||3 means 1 channel used for MNIST and 3 channels for SVHN). The symbol & is logical AND conjunction, meaning that both sizes are used (e.g. 10 & 32 means that there are two latent vectors one having 10 units and the other having 32 units, the former being explicit variables and the latter implicit).

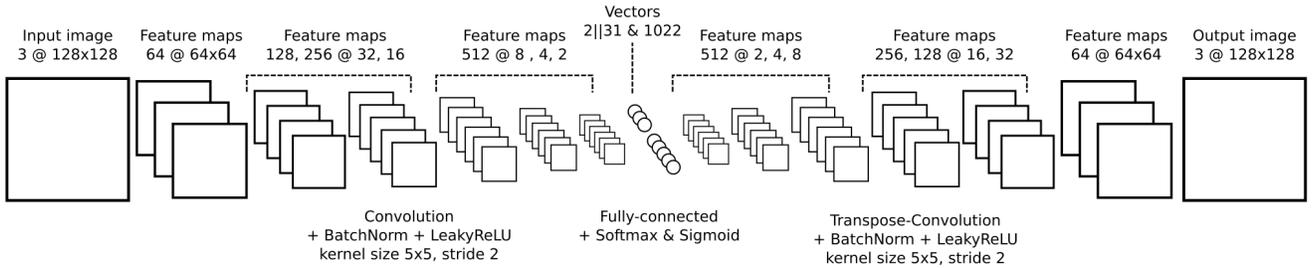

Figure 2: Neural network structure used in the CelebA, Cityscapes, and Chairs experiments. The notation represents: channels @ feature maps (or vectors) size. To avoid clutter the same label has been used for multiple maps (e.g. 128, 256 @ 32, 16 represents two sets, one with 128 maps of size $32 \times 32$, the other with 256 maps of size $16 \times 16$). The number of explicit units is 2 for all the datasets but the Chairs, which has 31 labels. The number of implicit units is 1022 for all the CelebA tests, 510 for Cityscape, and 481 for Chairs.

## B. Hyperparameters

| Dataset | Learning rate | $\lambda_i$ | $\lambda_e$ | # implicit | # explicit | mini-batch |
|---|---|---|---|---|---|---|
| MNIST | $10^{-4}$ | 0.5 | 1.0 | 32 | 10 | 128 |
| SVHN | $10^{-4}$ | 0.1 | 1.0 | 32 | 10 | 128 |
| Chairs | $10^{-5}$ | 0.1 | 1.0 | 481 | 31 | 64 |
| Cityscape | $10^{-5}$ | 0.1 | 1.0 | 510 | 2 | 64 |
| CelebA (male↔female) | $10^{-5}$ | 0.1 | 1.0 | 1022 | 2 | 32 |
| CelebA (glasses↔no-glasses) | $10^{-5}$ | 0.1 | 1.0 | 1022 | 2 | 32 |

Table 1: Hyperparameters used in the experiments. Learning rate of the Adam optimizer, implicit ($\lambda_i$) and explicit ($\lambda_e$) mixing coefficient values, implicit and explicit number of units used in the code, mini-batch size at training time.



\end{document}